\documentclass[letterpaper]{article} 
\usepackage{thesty} 
\usepackage{times}  
\usepackage{helvet}  
\usepackage{courier}  
\usepackage[hyphens]{url}  
\usepackage{graphicx} 
\usepackage{natbib}  
\usepackage{caption} 
\usepackage{algorithm}
\usepackage{algorithmic}
\usepackage{tabularx}
\usepackage{booktabs}
\usepackage{multirow}
\usepackage{multicol}
\usepackage{amsmath}
\usepackage{longtable}

\usepackage{newfloat}
\usepackage{listings}

\DeclareCaptionStyle{ruled}{labelfont=normalfont,labelsep=colon,strut=off} 
\floatstyle{ruled}
\newfloat{listing}{tb}{lst}{}
\floatname{listing}{Listing}

\title{A Unified Multi-Agent Framework for Universal Multimodal \\ Understanding and Generation}

\author {
    Jiulin Li\textsuperscript{\rm1,2},
    Ping Huang\textsuperscript{\rm1$\dagger$},
    Yexin Li\textsuperscript{\rm1},
    Shuo Chen\textsuperscript{\rm1$\dagger$},\\
    Juewen Hu\textsuperscript{\rm1},
    Ye Tian\textsuperscript{\rm2},
    \vspace{0.1cm}
}

\affiliations {
    \textsuperscript{\rm 1} State Key Laboratory of General Artificial Intelligence, BIGAI\\
    \textsuperscript{\rm 2} State Key Laboratory of Switching and Networking, Beijing University of Posts and Telecommunications\\
    \vspace{0.1cm}
    \{huangping, chenshuo\}@bigai.ai
    \vspace{0.1cm}
}

\begin{document}

\maketitle

\begingroup
  \renewcommand\thefootnote{*}
  \footnotetext{$\dagger$ denotes corresponding authors.}
\endgroup

\begin{abstract}
Real-world multimodal applications often require any-to-any capabilities, enabling both understanding and generation across modalities including text, image, audio, and video. However, integrating the strengths of autoregressive language models (LLMs) for reasoning and diffusion models for high-fidelity generation remains challenging. Existing approaches rely on rigid pipelines or tightly coupled architectures, limiting flexibility and scalability. We propose MAGUS (Multi-Agent Guided Unified Multimodal System), a modular framework that unifies multimodal understanding and generation via two decoupled phases: Cognition and Deliberation. MAGUS enables symbolic multi-agent collaboration within a shared textual workspace. In the Cognition phase, three role-conditioned multimodal LLM agents—\textit{Perceiver}, \textit{Planner}, and \textit{Reflector}—engage in collaborative dialogue to perform structured understanding and planning. The Deliberation phase incorporates a Growth-Aware Search mechanism that orchestrates LLM-based reasoning and diffusion-based generation in a mutually reinforcing manner. MAGUS supports plug-and-play extensibility, scalable any-to-any modality conversion, and semantic alignment—all without the need for joint training. Experiments across multiple benchmarks, including image, video, and audio generation, as well as cross-modal instruction following, demonstrate that MAGUS outperforms strong baselines and state-of-the-art systems. Notably, on the MME benchmark, MAGUS surpasses the powerful closed-source model GPT-4o. 

\end{abstract}
\begin{figure}[!htb]
\centering
\includegraphics[width=\columnwidth]{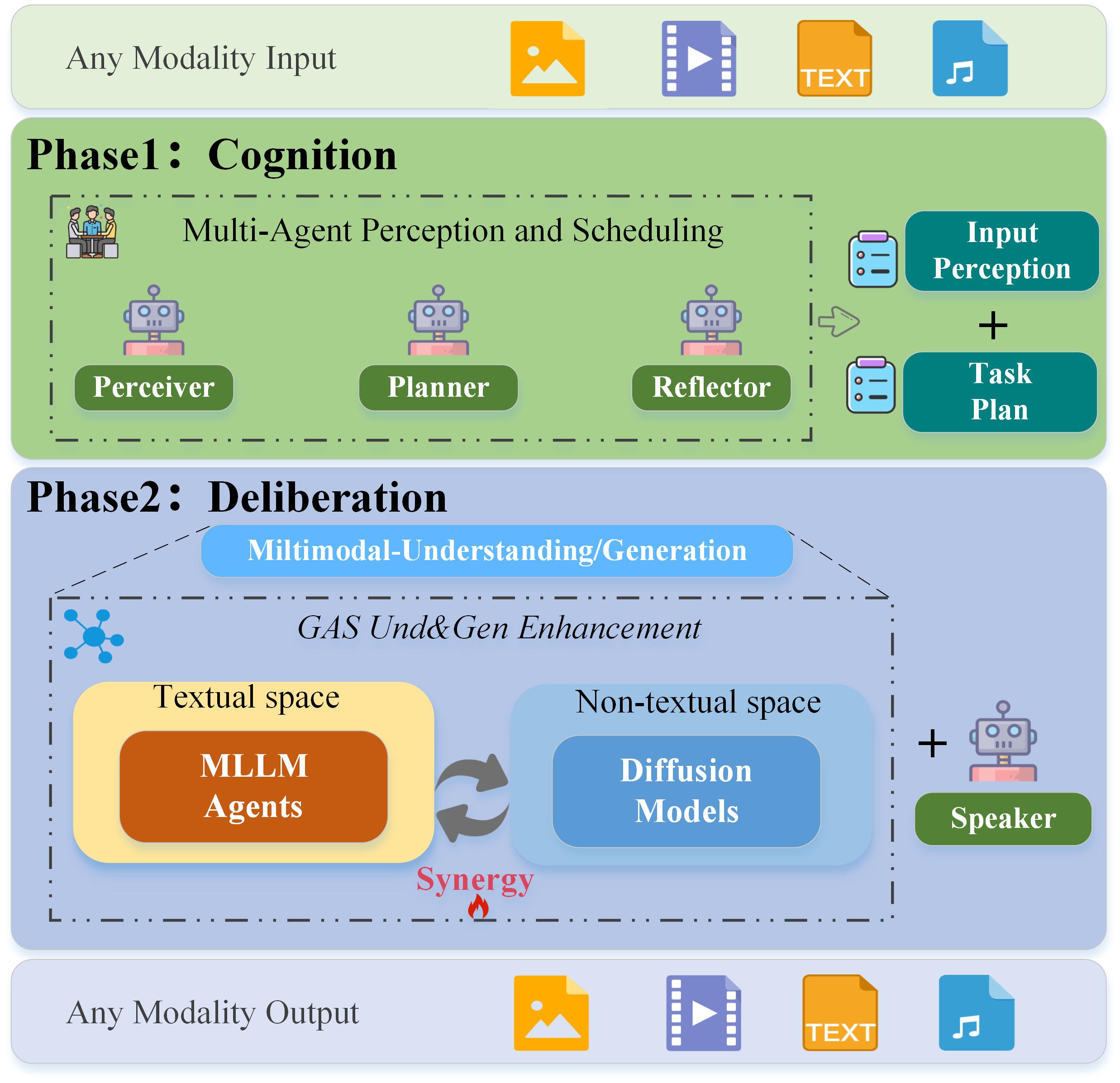} 
\caption{\textbf{Overview of the MAGUS framework}, which decomposes multimodal reasoning and generation into two cooperative phases—Cognition and Deliberation—via modular agents operating across textual and non-textual spaces.}
\label{fig1}
\end{figure}

\section{Introduction} 

Large language models (LLMs) have achieved remarkable success in natural language understanding and generation. This progress has led to the emergence of multimodal large language models (MLLMs)~\cite{bai2025qwen2,xu2025qwen2,liao2025mogao}, which leverage language as a unified interface for cross-modal reasoning. In parallel, diffusion models have become the dominant approach for high-fidelity generation in visual~\cite{wan2025wan,zheng2024open,esser2024scaling} and audio domains~\cite{liu2023audioldm,evans2024fast}, offering an alternative to the autoregressive paradigm adopted by LLMs.

Real-world multimodal tasks often require any-to-any capabilities, where models can process any modality as input and generate any modality as output—e.g., audio-to-image or text-to-video—spanning both understanding and generation across modalities. As such, the paradigm gap between autoregressive LLMs (strong in reasoning and semantics) and diffusion models (strong in fidelity and generation) becomes a key obstacle. This leads to the central challenge: how to build a unified multimodal framework that supports flexible any-to-any tasks, while integrating the complementary strengths of LLMs and diffusion models. 

Current approaches fall into two categories: modular pipelines that chain pretrained models~\cite{lai2024spider,wu2024next}, and end-to-end unified architectures~\cite{li2025dual,liao2025mogao,deng2025emerging}. The former often lacks tight integration for coherent reasoning, while the latter requires costly joint training, sacrifices modularity, and still falls short of fully supporting general-purpose multimodal generation. Both approaches face challenges in extending to new modalities and provide limited flexibility for model reuse and upgrading. This underscores the need for a framework that is both unified in control and modular in design—enabling scalable, interpretable, and composable multimodal intelligence.


In this work, we introduce MAGUS (Multi-Agent Guided Unified Multimodal System), a novel framework that unifies multimodal understanding and generation through a modular, multi-agent architecture inspired by the Global Workspace Theory~\cite{baars1993cognitive}. MAGUS decouples multimodal processing into two distinct phases—Cognition and Deliberation—and instantiates a multi-agent system within an MLLM through symbolic role-switching. During the Cognition phase, agents such as the \textit{Perceiver}, \textit{Planner}, and \textit{Reflector} collaborate within a shared textual workspace to analyze inputs, infer goals, and plan tasks. In the Deliberation phase, MAGUS employs a novel Growth-Aware Search mechanism to generate modality-specific outputs, orchestrating LLM-based reasoning and diffusion-based generation in a mutually reinforcing manner.

Unlike prior monolithic architectures that tightly entangle LLMs and diffusion models, MAGUS adopts a decoupled yet synergistic design—leveraging MLLMs for semantics and reasoning, and diffusion models for high-fidelity, modality-specific generation. All coordination and control occur entirely within the textual space, enabling seamless integration of state-of-the-art MLLMs and generative models without requiring joint training. MAGUS also supports flexible module replacement and upgrading, thereby enhancing scalability and modularity.

Importantly, MAGUS goes beyond simple model composition by enabling its components to interact and reinforce one another through a shared semantic space, resulting in significant gains in both multimodal understanding and generation. It not only outperforms its individual base models but also surpasses many state-of-the-art systems. Evaluated on our proposed MM-Instruction-Test benchmark, MAGUS demonstrates strong cross-modal instruction-following capabilities. It effectively handles complex tasks—including instruction execution, semantic-guided generation, and goal-directed synthesis—many of which remain challenging.


Our contributions are summarized as follows:

\begin{itemize}
    \item  We propose MAGUS, a novel decoupled two-phase multi-agent framework for general-purpose multimodal understanding and generation. Built around a unified multimodal LLM, MAGUS enables interpretable reasoning and modular, plug-and-play integration of pretrained models—without requiring joint retraining.

    \item  We introduce Growth-Aware Search (GAS), an agent-based search algorithm in MAGUS that leverages iterative rollouts and feedback loops between the MLLM and diffusion models to jointly improve multimodal understanding and generation.

    \item We show that MAGUS outperforms its base models and state-of-the-art methods in multimodal understanding and generation, with strong instruction-following capabilities. We introduce MM-Instruction-Test, a compact benchmark for evaluating such abilities.
\end{itemize}

\section{Related work}



\subsection{Modality Extension for LLMs}

A common architecture for multimodal LLMs uses modality-specific encoders with lightweight projectors to align features into the language space~\cite{wu2024next,lai2024spider}. Early methods typically freeze the language model and train only the projectors, enabling efficient adaptation but often leading to limited capacity and semantic misalignment. Recent models like VITA~\cite{fu2024vita}, Qwen-VL~\cite{xu2025qwen2}, Qwen-Omni~\cite{bai2025qwen2} improve performance by jointly pretraining encoders and projectors with the LLM. However, this tight coupling reduces modularity and makes it costly to extend generation to new modalities.

Building on this paradigm, MAGUS adopts a strong unified MLLM foundation (e.g., Qwen-Omni) to retain powerful multimodal understanding. It introduces a shared semantic space that bridges non-textual inputs to the language domain, enabling expressive generation. MAGUS allows plug-and-play integration of modality-specific generators without retraining the core model, achieving scalable and flexible multimodal expression. This design enables comprehensive extension of both understanding and generation across all modalities. 


\subsection{Perception and Synthesis}
While LLMs~\cite{achiam2023gpt} unify understanding and generation within text, extending them to perceptual modalities (e.g., image, video, audio) presents challenges—particularly in high-fidelity synthesis. One line of work, such as NextGPT~\cite{wu2024next} and Spider~\cite{lai2024spider}, links LLMs with diffusion models via projection layers, allowing the LLM to control generation. However, these systems often rely on task-specific tuning and exhibit shallow semantic coordination, limiting their generalization and reasoning capabilities. Another line, including BAGEL~\cite{deng2025emerging}, Mogao~\cite{liao2025mogao}, Dual-Diffusion~\cite{li2025dual}, and Vila-U~\cite{wu2024vila}, explores unified multimodal architectures to support generation across text and image. While this improves cross-modal alignment, the model is still unable to generate video and remains far behind advanced diffusion models~\cite{wan2025wan,jiang2025vace} in terms of generative capability and modality coverage. 

In contrast, MAGUS leverages a strong MLLM for holistic multimodal understanding, supports plug-and-play integration with advanced video generators, and enables bidirectional improvement via GAS. Its two-stage design further ensures excellent instruction following and task execution. 

\begin{figure*}[!htb]
\centering
\includegraphics[width=\textwidth]{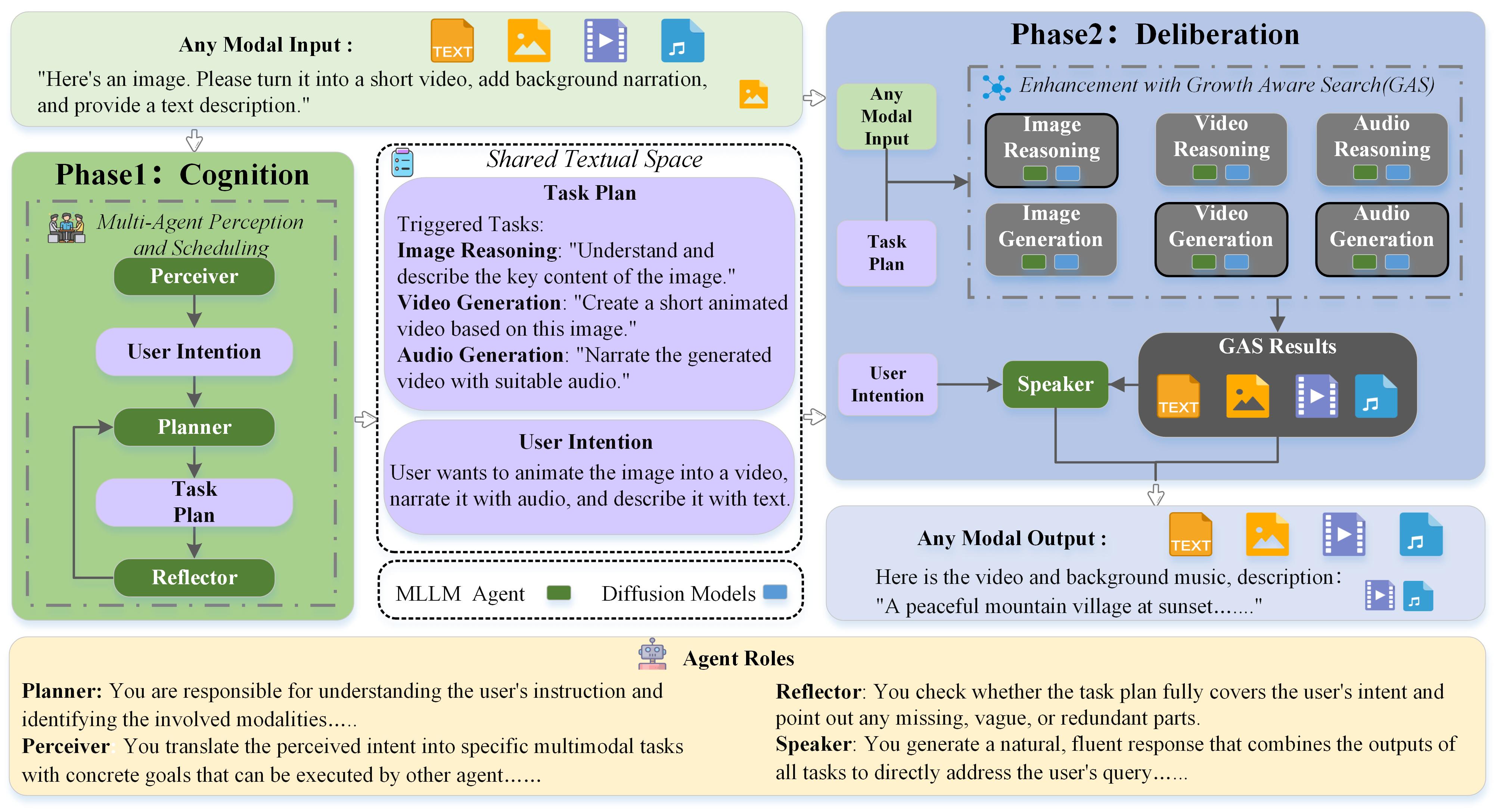} 
\caption{
        \textbf{Overview of the MANGUS.}
MAGUS is built on a unified multimodal LLM (MLLM) for perception and reasoning, paired with task-specific diffusion models for generation. Agents are lightweight, role-specialized variants of the MLLM, enabling training-free, interpretable coordination. The framework separates perception from deliberation. Final outputs are aggregated and verbalized by the Speaker agent, producing coherent multimodal responses.
}
\label{fig2}
\end{figure*}

\subsection{Multi-Agent Systems}
Multi-agent frameworks have emerged to handle the growing complexity of multimodal tasks. MM-ReAct~\cite{yang2023mm} and ToolLLM~\cite{qin2023toolllm} use language models to coordinate external tools, while AudioGPT~\cite{huang2024audiogpt} and WavCraft~\cite{liang2024wavcraft} connect multiple foundation models for audio-related tasks, and the VideoRefer~\cite{yuan2025videorefer} focuses on visual tasks. However, these systems often rely on large closed-source models and are limited to specific modalities or functions. In contrast, MAGUS introduces a compact and unified multi-agent framework that supports both understanding and generation for all modalities(Text,Image,Video,Audio).

\section{Methods}
As illustrated in Figure~\ref{fig2}, we propose MAGUS, a two-stage framework for unified multimodal reasoning and generation. Inspired by the Global Workspace Theory (GWT)~\cite{baars1993cognitive}, MAGUS separates \emph{Cognition} and \emph{Deliberation}, echoing the cognitive division between sensing and reasoning. In the Cognition stage, expert agents collaboratively process inputs and formulate tasks. The Deliberation stage then employs a Growth-Aware Search (GAS) mechanism to execute these tasks by dynamically invoking MLLMs and diffusion models for cross-modal reasoning and high-fidelity generation—without requiring joint training.

\subsection{Phase 1: Cognition}
Unified multimodal tasks often involve complex semantics, latent user intent, and cross-modal references, which cannot be fully resolved by single-pass inference. MAGUS addresses this by introducing an explicit Cognition phase, where the system deeply interprets user instructions and decomposes them into structured, modality-aware task plans. This phase simulates cognitive preprocessing: it contextualizes user goals, grounds them in multimodal inputs, and formulates executable actions for the downstream generation. The output includes both the user's high-level intent and a stepwise plan describing what to generate, understand, or retrieve—and with which modality—serving as a precise blueprint for the next stage.

\begin{figure*}[!htb]
\centering
\includegraphics[width=\textwidth]{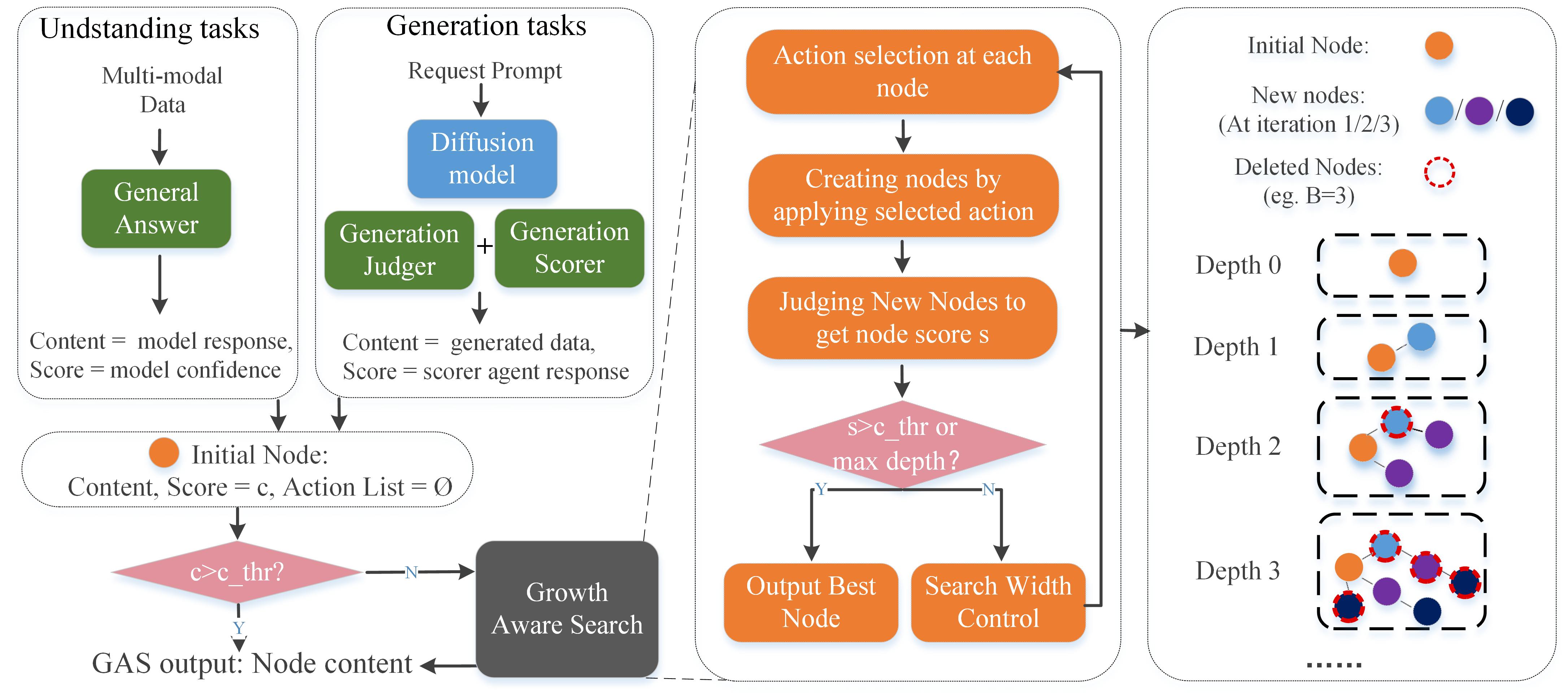} 
\caption{
        \textbf{Overview of the proposed Growth-Aware Search (GAS) mechanism for cross-modal task enhancement.} 
Given an initial understanding or generation result, GAS incrementally applies expert actions, scores new hypotheses, and iteratively searches for optimal content using confidence-guided breadth control and early stopping.
}
\label{fig3}
\end{figure*}

\subsubsection{Multi-Agent Cognition and Scheduling}

The Cognition phase is realized through a collaborative dialogue among three role-conditioned MLLM agents: \textit{Perceiver}, \textit{Planner}, and \textit{Reflector}. The \textit{Perceiver} first interprets the user prompt and any accompanying multimodal context, producing a concise semantic representation of the task. Next, the \textit{Planner} constructs a structured task plan, specifying which modality-specific operations (e.g., image generation, audio reasoning) need to be performed. The \textit{Reflector} then evaluates the proposed plan against the inferred user intent, identifying missing or redundant steps and initiating plan revisions when necessary. This multi-round refinement ensures that the task plan is accurate, complete, and executable. Once validated, both the plan and intent representation are passed to the Deliberation phase for execution. 

\subsection{Phase 2: Deliberation}

The Deliberation phase is responsible for executing the structured task plan generated in the Cognition stage and performing corresponding multimodal reasoning and generation tasks. Depending on the task type—e.g., image reasoning, generation, video understanding, or audio generation—MAGUS selectively activates the corresponding tasks' execution. 

Rather than relying on task-specific pipelines, we propose Growth-Aware Search (GAS), a unified, training-free mechanism that enables dynamic, bidirectional enhancement between multimodal reasoning and generation. GAS allows MLLMs and diffusion models to mutually refine each other, going beyond sequential execution. Once refined, the outputs—combined with the user intent from the cognition stage—are passed to the \textit{Speaker Agent}, which generates a coherent, query-aligned natural language response. 

\subsubsection{Growth-Aware Search for Multi-modal Reasoning and Generation Enhancement}
As illustrated in Figure~\ref{fig3}, GAS operates as a guided, constrained action selection process, driven by confidence-based scoring and dynamic collaboration among agents.

\textbf{Initialization.} GAS starts with a coarse initial attempt to solve the task. For understanding tasks, a \textit{Answer Agent} directly produces a response and computes a confidence score $s_u$ by averaging the token-level probabilities from the language model output:

$$
s_u = \frac{1}{T} \sum_{t=1}^{T} P(y_t \mid y_{<t}, x)
$$
where $y_t$ is the $t$-th token of the output and $x$ is the input. For generation tasks, a diffusion model generates sample $d$, which is evaluated cascadedly: \textit{Judger} evaluates $d$, from multiple perspectives, such as semantic alignment with the instruction, quality, and coherence, and generates the judgment text $j_g$. Subsequently, the \textit{Scorer} assesses quality based on $j_g$. The final confidence score is:
$$
s_g = \mathrm{Scorer}(\mathrm{Judger}(d))
$$

If the corresponding confidence score $s_u$ or $s_g$ exceeds a predefined threshold $c_{\text{thr}}$, the result is accepted. Otherwise, the system triggers the refinement procedure.

\begin{figure*}[!htb]
\centering
\includegraphics[width=\textwidth]{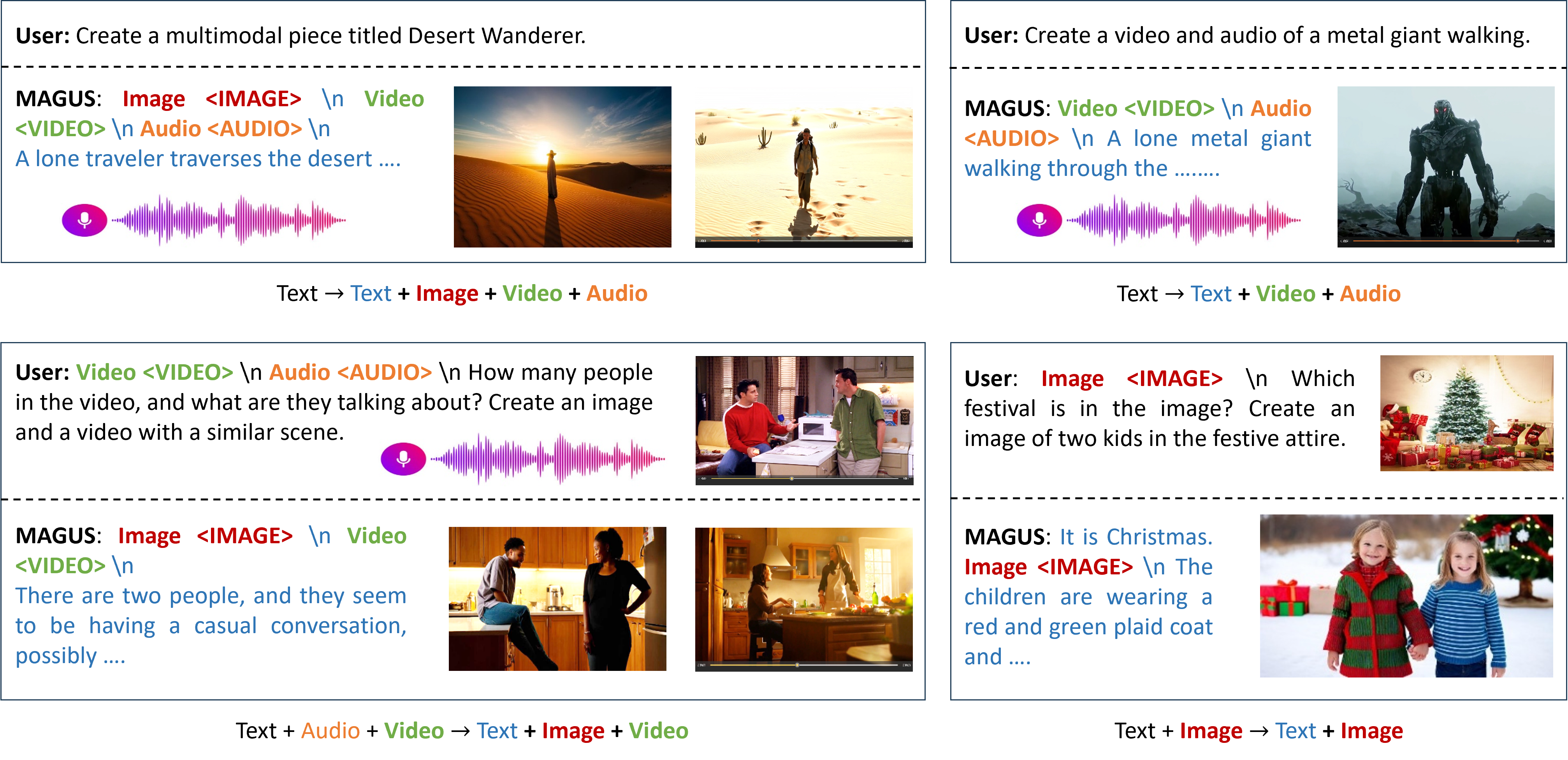} 
\caption{
        \textbf{Examples of unified multimodal understanding and generation by MAGUS. } }
\label{fig4}
\end{figure*}

\textbf{Node Expansion and Search Procedure.}
GAS represents each candidate solution as a node $n_i = (x_i, s_i, \mathcal{A}_i)$, where $x_i$ is the generated content (e.g., text, image), $s_i$ represents the node score obtained through the aforementioned scoring method, and $\mathcal{A}_i$ is the ordered list of actions leading to $x_i$. The search starts from the initial node $n_0$ and proceeds via a depth- and width-constrained expansion to control the search complexity.

Formally, GAS maintains a beam of up to $B$ candidate nodes. At each iteration, every node in the current beam is considered for expansion. For each node $n_i$, the \textit{Action Selector} proposes one action $a \in \mathcal{A}_{\text{space}} \setminus \mathcal{A}_i$, where $\mathcal{A}_{\text{space}}$ is the action space. The selected action is then applied to generate new successor nodes:
$$
n_j = \mathrm{Apply}(n_i, a) = \left(x_j, s_j, \mathcal{A}_i \cup \{a\}\right),
$$
where $ x_j $ is the new state, $ s_j $ is its score, and $ \mathcal{A}_i \cup \{a\} $ the updated action sequence. The global candidate pool—consisting of both existing and newly generated nodes—is then updated. The beam is refreshed by selecting the top-$B$ highest-scoring nodes from this pool, while lower-ranked candidates are pruned if the total exceeds $B$. This iterative process continues until one of the following termination conditions is met: (1) a node's score exceeds the confidence threshold $c_{\text{thr}}$; (2) the maximum search depth $D$ is reached; or (3) no further valid actions can be applied.


\textbf{Actions.}
GAS defines a domain-specific action space $\mathcal{A}_{\text{space}}$ tailored to the task type. For example, in understanding tasks, two categories of actions are supported: (1) Expert Interpretation Actions: A set of task-specific agents (e.g., cultural analysts, logical reasoning agents, visual experts) process the multimodal input to produce side-perspective insights. These outputs are appended to the current node content as auxiliary references. (2) Generative Augmentation Actions: A diffusion model is used to synthesize auxiliary content (e.g., high-resolution reconstructions) from the original multimodal input. The resulting data enhances perceptual depth and inference reliability. Applying an action to a node augments its content with new information. A \textit{Summarization Agent} then processes the enriched content to generate an updated understanding response $x_j$, which is subsequently scored to yield $s_j$. The new node $n_j$ is constructed by appending the applied action to the parent’s history.

In contrast, for generation tasks, GAS employs only task-specific agents as action primitives. Each action modifies the prompt for diffusion models(e.g., via prompt refinement or constraint injection), based on both the content and the judgment $j_g$ of the current node, and is executed by specialized prompt experts. The modified prompt is passed to a diffusion model to produce new multimodal content, which is then evaluated by the Judger–Scorer pair to assign a score.

This unified yet flexible formulation allows GAS to iteratively interleave symbolic reasoning and sub-symbolic generation within a shared search paradigm, enabling them to mutually reinforce each other. Implementation details for each task, along with the definitions and workflows of the actions, are provided in the appendix.

\textbf{Output and Generality.}
The highest-confidence node is returned as the final result. This procedure enhances LLM outputs with symbolic multi-hop reasoning and refines diffusion outputs with task-specific evaluation loops—bridging understanding and generation through coordinated agent interaction. GAS is easily extensible: new modalities or tools can be added as new actions without retraining, making the system adaptable to evolving task demands.


\section{Experiments}
\subsection{Implatention Details}
\subsubsection{Models}
In our experiments, we selected several state-of-the-art models to demonstrate the strong adaptability of our framework. For the understanding module, we used Qwen2.5-Omni 7B~\cite{xu2025qwen2} as the backbone model, which possesses both language capabilities and full-modal perception abilities. The generation module consists of the Wan-VACE 1.3B~\cite{jiang2025vace} model for video and image generation, capable of producing both images and videos, as well as the audioldm-s-full-v2~\cite{liu2023audioldm} model for audio generation. 

\subsubsection{Hyperparameter Configuration}
For the key hyperparameter $c_{\text{thr}}$ in the GAS algorithm, we report the optimal value based on performance on understanding tasks, with a detailed analysis presented in Section~\ref{sec:cablation}. For generation tasks, due to the high computational cost, we adopt a fixed value of $c_{\text{thr}} = 0.6$ without further hyperparameter tuning. All generation-related parameters for the diffusion models are provided in the Appendix. 

\subsubsection{Dataset and Metrics}
For multimodal understanding tasks, we evaluated our framework on MME~\cite{fu2023mme}, MMAU~\cite{sakshi2024mmau}, and VideoEspresso~\cite{han2025videoespresso} datasets, reporting results according to their respective metrics. For generation tasks, we tested on Geneval~\cite{ghosh2023geneval}, VBench~\cite{huang2024vbench}, and AudioCaps~\cite{kim2019audiocaps} datasets. Specifically, we report the corresponding benchmark metrics score for Geneval and VBench. For AudioCaps, we use the Audio Aesthetics Score (AES)~\cite{tjandra2025meta}—focusing on Production Quality (PQ) and Production Complexity (PC)—and the Fréchet Distance (FD) to assess generated audio quality and distributional divergence.

\subsection{Any-to-Any Understanding and Generation}


As illustrated in Figure~\ref{fig4}, our MAGUS framework enables flexible any-to-any modality conversion while maintaining robust instruction comprehension. The system demonstrates comprehensive multimodal capabilities through unified processing of diverse input-output combinations, including complex cross-modal tasks. Quantitative results in the following section confirm that MAGUS achieves consistent performance across all modality conversions without specialized tuning. Quantitative results in the following section confirm that MAGUS achieves consistent performance across all modality conversions without specialized tuning. All experiments are conducted by directly applying the corresponding tasks to MAGUS’s GAS pipeline.

\subsubsection{Multimodal Understanding}
\begin{table}[!htb]
\centering
\begin{tabular}{lccc}
\toprule
Model & MME-P & MME-C & MME-Sum \\
\midrule
VILA-U-7B & 1402 &-- & --\\
Janus-Pro-7B & 1567 & -- & --  \\
Mogao-7B & 1592 & -- & --  \\
VITA & -- & -- & 2097 \\
Gemini-1.5-pro &--	&-- & 2111 \\
GPT-4o & -- & -- & 2310\\
Qwen2.5-Omni-7B  & 1545 & 607 & 2155 \\
MAGUS(ours) & \textbf{1648} & \textbf{674} & \textbf{2322} \\
\bottomrule
\end{tabular}
\caption{Performance Comparison on MME Benchmark. Results are averaged over 5 runs with negligible variance ($<$0.1). MAGUS(base MLLM: Qwen2.5-Omni-7B).}
\label{tab:mme_results}
\end{table}

\begin{table}[!htb]

\centering
\begin{tabular}{lc}
\toprule
Model & Total score \\
\midrule
Gemini-1.5-pro	 & 44.2 \\
Kangaroo-8B & 44.1 \\
Qwen2.5-Omni-7B & 53.2 \\
MAGUS(ours) & \textbf{53.3} \\
\bottomrule
\end{tabular}
\caption{Performance Comparison on VideoEspresso Benchmark. Results are averaged over 5 runs with negligible variance ($<$0.1). MAGUS(MLLM: Qwen2.5-Omni-7B).}
\label{tab:video_results}
\end{table}

\begin{table}[!htb]

\centering
\begin{tabular}{lcccc}
\toprule
Model & Sound & Music & Speech & Sum \\
\midrule
Qwen-Omni-7B & 41.1 & 39.8 & 49.3 & 43.4 \\
MAGUS & \textbf{71.8} & \textbf{57.2} & \textbf{58.6} & \textbf{61.7} \\
\bottomrule
\end{tabular}
\caption{Audio Reasoning on MMAU test-mini-split. Results are averaged over 5 runs with negligible variance ($<$0.1). MAGUS(MLLM: Qwen2.5-Omni-7B)}
\label{tab:audio_results}
\end{table}

Table~\ref{tab:mme_results} shows the performance of MAGUS on the MME benchmark, where it achieves the highest scores across all reported metrics, outperforming a strong closed-source model GPT-4o~\cite{hurst2024gpt}. This demonstrates MAGUS’s superior multimodal understanding capability. On the VideoEspresso benchmark (Table~\ref{tab:video_results}), MAGUS slightly surpasses Qwen2.5-Omni-7B and other competitive models, indicating its effectiveness in video reasoning tasks. Table~\ref{tab:audio_results} highlights MAGUS’s substantial gains in audio reasoning, with significant improvements over Qwen-Omni-7B across sound, music, speech, and overall scores. This validates the framework’s strength in handling diverse audio modalities. Across all benchmarks, results are averaged over multiple runs with negligible variance, confirming the robustness and consistency of MAGUS.

\subsubsection{Multimodal Generation}

\begin{table}[!htb]
\centering
\begin{tabular}{lc}
\toprule
Model & Total Score \\
\midrule
Wan-VACE-1.3B & 37.6 \\
+LLM Extend Prmopt  & 67.7 \\
MAGUS                    & \textbf{71.1} \\
\bottomrule
\end{tabular}
\caption{Image Generation Results on GenEval Benchmark. MAGUS (Image generator: Wan-VACE-1.3B)}
\label{tab:generation_geneval}
\end{table}

\begin{table}[!htb]
\centering
\begin{tabular}{lccc}
\toprule
Model & Quality & Semitic & Total \\
\midrule
Wan-VACE-1.3B & 80.1 & 66.1 & 77.3 \\
+LLM Extend Prompt  & \textbf{81.4} & 73.5 & 79.8 \\
MAGUS             & 81.0 & \textbf{77.9} & \textbf{80.4} \\
\bottomrule
\end{tabular}
\caption{Video Generation Results on VBench Benchmark. MAGUS (Video generator: Wan-VACE-1.3B)}
\label{tab:generation_vbench}
\end{table}

\begin{table}[!htb]
\centering
\begin{tabular}{lccc}
\toprule
Model & PC & PQ & FD\\
\midrule
Audioldm-s-full-v2 & 3.1 & \textbf{5.8} & \textbf{7.7}  \\
+LLM Extend Prompt       & 3.1 & 5.3 & 9.9  \\
MAGUS                    & \textbf{3.2} & 5.4 & 9.3  \\
\bottomrule
\end{tabular}
\caption{Audio Generation Results on AudioCaps. MAGUS(Audio generator:Audioldm-s-full-v2)}
\label{tab:generation_audiocaps}
\end{table}

Table~\ref{tab:generation_geneval} reports image generation results on the GenEval benchmark, where MAGUS significantly outperforms the baseline Wan-VACE-1.3B and its LLM prompt-extended variant, achieving the highest total score. This highlights MAGUS’s superior capability in integrating language guidance for enhanced image synthesis. On the VBench video generation benchmark (Table~\ref{tab:generation_vbench}), MAGUS attains the best overall total score, notably improving semantic consistency while maintaining competitive generation quality compared to the baseline and prompt-extended models. Table~\ref{tab:generation_audiocaps} presents audio generation results on AudioCaps. MAGUS achieves the highest Perceptual Coverage (PC) and a balanced performance in Perceptual Quality (PQ) and Frechet Distance (FD), demonstrating robust audio synthesis capabilities without sacrificing fidelity. Overall, MAGUS consistently advances multimodal generation quality across image, video, and audio domains, validating its effectiveness as a unified generation framework built upon strong modality-specific backbones.

\subsection{Instruction Following Evaluation}

\begin{table}[H]
\centering
\begin{tabular}{lc}
\toprule
\textbf{Metric} & \textbf{MAGUS} \\
\midrule
Strict Match Accuracy (\%) & 75.0 \\
Flexible Coverage Accuracy (\%) & 90.0 \\
\bottomrule
\end{tabular}
\caption{Accuracy on MM-Instruction-Test}
\label{tab:instruction_following}
\end{table}

We created a compact benchmark called MM-Instruction-Test, covering multimodal generation instruction-following tasks from unimodal to quadrimodal generation. Details are provided in the Appendix. We evaluate MAGUS on MM-Instruction-Test using qualitative examples and expert ratings. In Table~\ref{tab:instruction_following}, MAGUS achieves 75.0\% strict match accuracy and 90.0\% under the more lenient Flexible Coverage metric, reflecting a balance between precise adherence and modality coverage in multimodal generation.

\subsection{Confidence Threshold vs. Performance}
\label{sec:cablation}
\begin{figure}[!htb]
\centering
\includegraphics[width=\columnwidth]{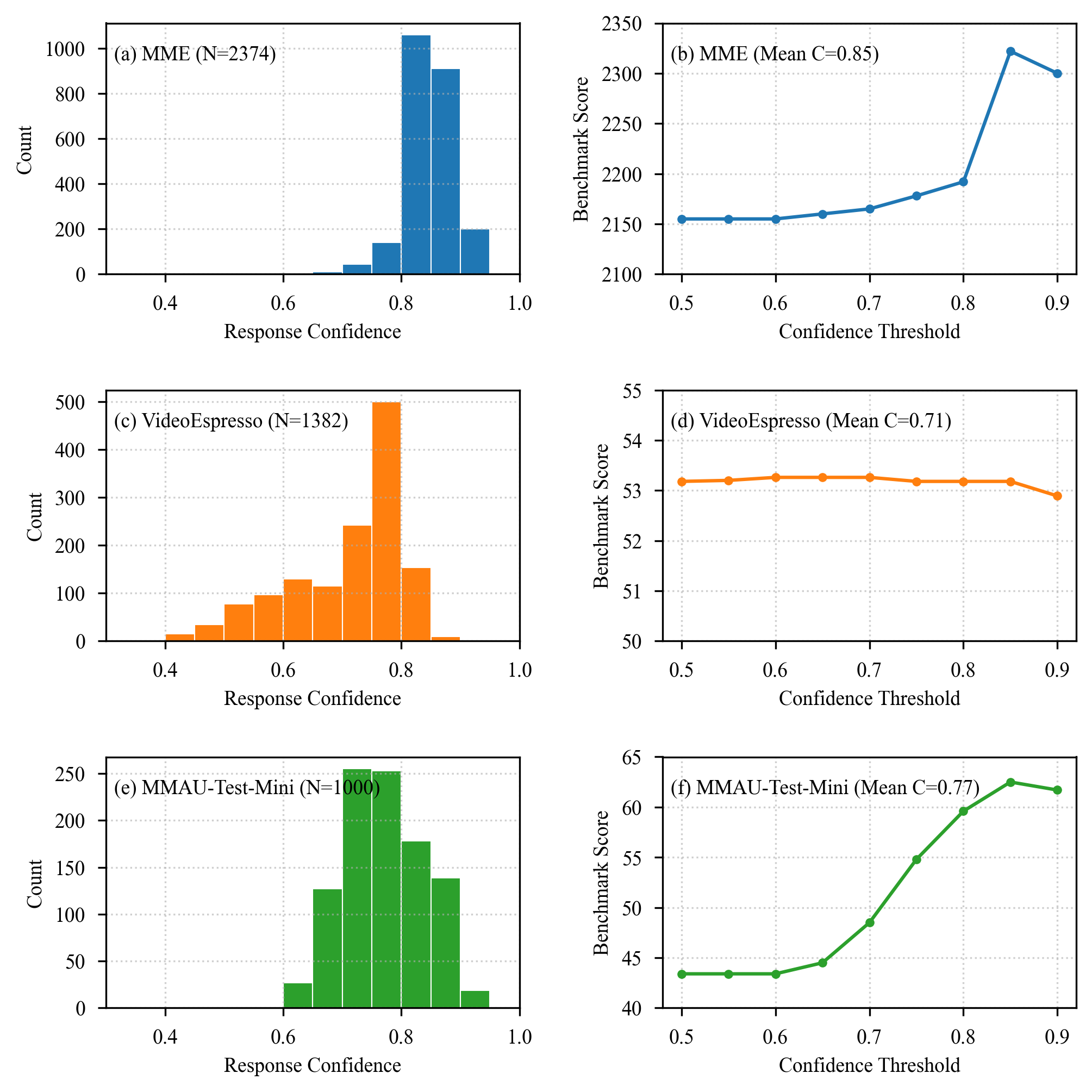} 
\caption{
\textbf{Confidence analysis on multimodal reasoning datasets.} Confidence distributions and threshold effects on model performance.
}
\label{fig5}
\end{figure}

We further investigate the impact of confidence thresholds on the overall performance of reasoning tasks, as shown in Figure~\ref{fig5}. For each modality, we visualize the distribution of prediction confidence scores and evaluate system performance under varying threshold settings. Results show that increasing the confidence threshold improves benchmark scores on MME and MMAU. When the threshold is low, most responses bypass the GAS process, resulting in performance comparable to the baseline. A performance drop at higher thresholds may be attributed to excessive reliance on the GAS mechanism, where all actions are triggered, and suboptimal experts may introduce hallucinated outputs. In contrast, VideoEspresso exhibits minimal variation, possibly due to the inherently complex nature of video reasoning tasks—expert recommendations tend to converge, and the auxiliary visual outputs offer limited additional benefit.

\section{Discussion}

This work does not aim to outperform all existing methods on every benchmark. Instead, we address two fundamental challenges in multimodal AI: (1) integrating autoregressive and diffusion-based models, and (2) unifying understanding and generation within a single framework. Rather than enforcing a shared representation space, MAGUS leverages the complementary strengths of specialized models through modular collaboration. 

The MAGUS framework benefits from strong modular decoupling, enabling seamless integration of evolving foundation models—such as LLMs and diffusion models—without costly joint retraining. The generative module can be flexibly replaced or specialized for target domains, supporting deployment of high-capacity or task-specific models (e.g., WaN 2.1 14B~\cite{wan2025wan}).

Future work can improve agent collaboration by enhancing coordination efficiency and reducing redundancy, further boosting system performance and scalability.

\section{Conclusion}

This work addresses the challenge of building a unified system for general-purpose multimodal understanding and generation. We present MAGUS, a modular and decoupled framework that separates cognition and expression into two explicit phases. By leveraging a shared language-centric semantic space, MAGUS bridges the gap between autoregressive reasoning and diffusion-based synthesis, enabling flexible collaboration between symbolic agents and modality-native generators. The Growth Aware Search (GAS) algorithm in MAGUS enables iterative refinement of the performance of MLLM and diffusion models in multimodal tasks. Empirical results show that MAGUS outperforms baselines in both understanding and generation benchmarks, achieving superior performance, fidelity, and controllability. MAGUS offers a practical and extensible path toward general-purpose multimodal intelligence by unifying reasoning and synthesis in a lightweight, agent-driven architecture.

\bibliography{thebib}

\newpage
\clearpage
\setcounter{section}{0}
\setcounter{figure}{0}
\setcounter{table}{0}
\setcounter{secnumdepth}{2}
\renewcommand{\thesection}{\arabic{section}}
\renewcommand{\thesubsection}{\thesection.\arabic{subsection}}

\section{Growth-Aware Search Details}

After entering the optimization phase of Growth-Aware Search (GAS), the algorithm employs a Selector to choose an action, which is then executed to generate a new node—along with its content and evaluation score. This section presents a formal description of how action selection and node generation are performed for both understanding and generation tasks.

\subsection{Node Information}

The structure of nodes differs slightly between reasoning and generation tasks. As shown in Figure~\ref{fig6}, a reasoning task node contains the following elements in its node content:

\begin{itemize}
    \item  Auxiliary Advice from Agents: Suggestions from auxiliary reasoning agents or auxiliary multimoda data from generation agents (empty in the initial node).
    \item  Original Input: The raw multimodal input (e.g., images, videos) along with a corresponding natural language question.
    \item Node Answer: The current response generated for the given input.
\end{itemize}

The score associated with a reasoning node is a scalar value, which is computed using the summarization agent by averaging the token-level probabilities from the language model output, as detailed in the main text.

For generation tasks, the node structure is illustrated in Figure~\ref{fig8}. Each generation node content includes:

\begin{itemize}
    \item Original Prompt: The initial text input used to guide generation.
    \item Node Prompt: A potentially refined version of the original prompt.
    \item Node Answer: The multimodal output generated from the node prompt by the diffusion model.
\end{itemize}

The generation node’s score consists of two components:

\begin{itemize}
    \item Score Value: A scalar representing the overall quality.
    \item Judgement: A textual assessment of the generated data, produced by the judger agent and used as input for scoring.
\end{itemize}

\subsection{Select Actions}

\begin{figure}[!htb]
\centering
\includegraphics[width=\columnwidth]{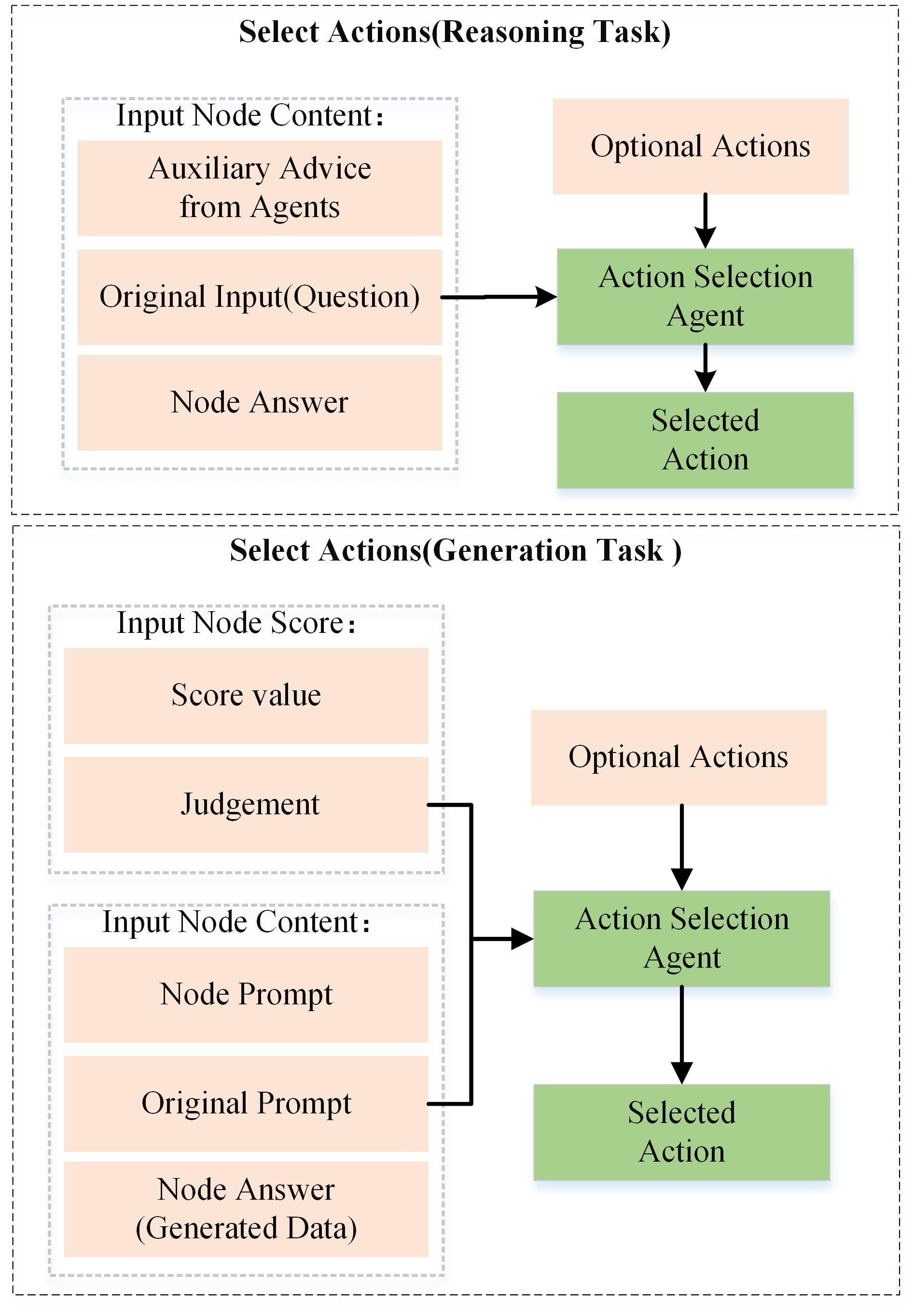} 
\caption{
\textbf{Action selection process for reasoning and generation tasks.}  Given current node content and scores, an Action Selection Agent selects the next optimal action from a set of options to guide subsequent reasoning or generation.
}
\label{fig6}
\end{figure}

\begin{figure}[!htb]
\centering
\includegraphics[width=\columnwidth]{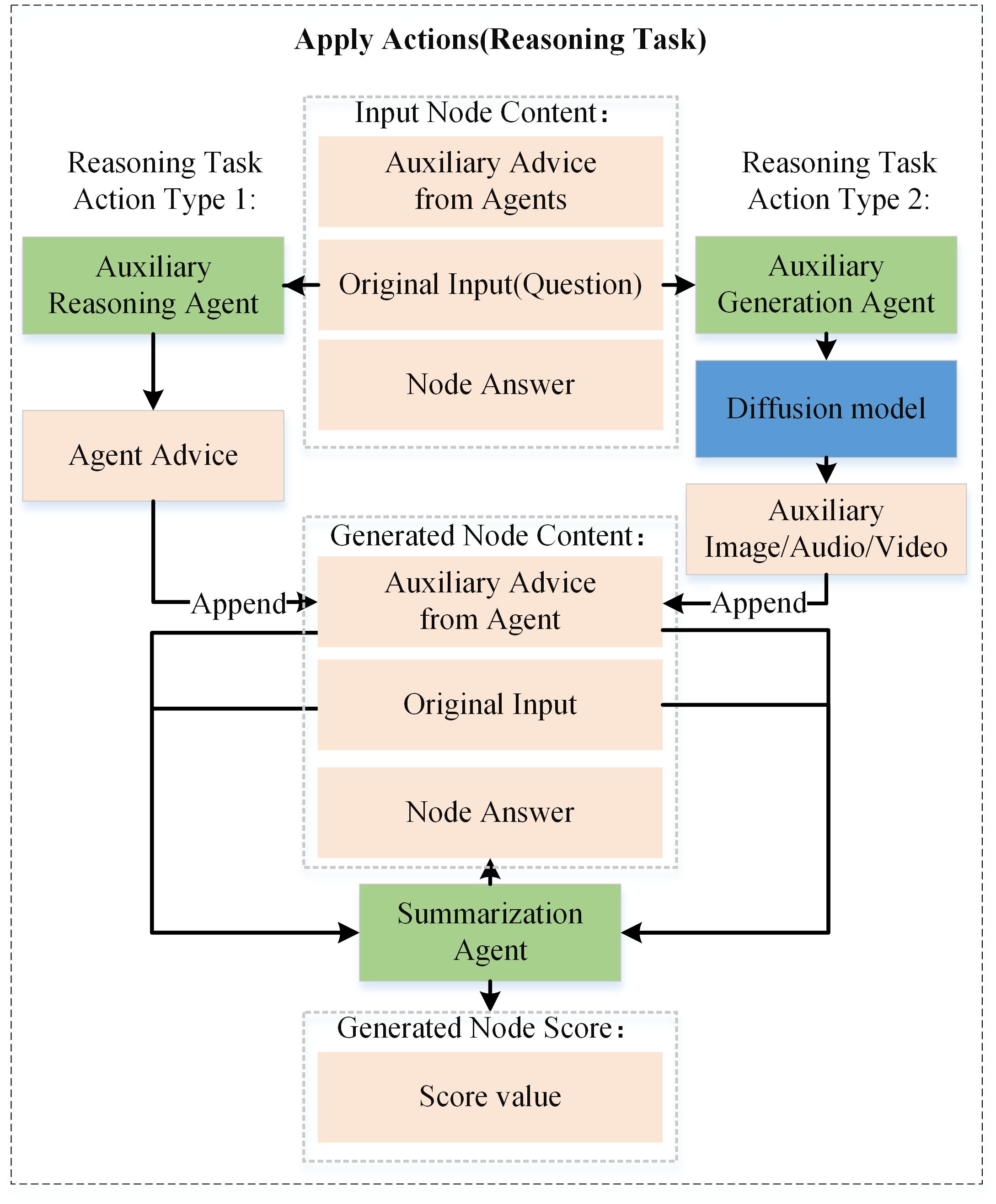} 
\caption{
\textbf{Workflow of applying actions in reasoning tasks.} MLLM agents provide auxiliary advice or generate intermediate multimodal content, which is evaluated by a summarization agent to assign a node score.
}
\label{fig7}
\end{figure}

\begin{figure}[!htb]
\centering
\includegraphics[width=\columnwidth]{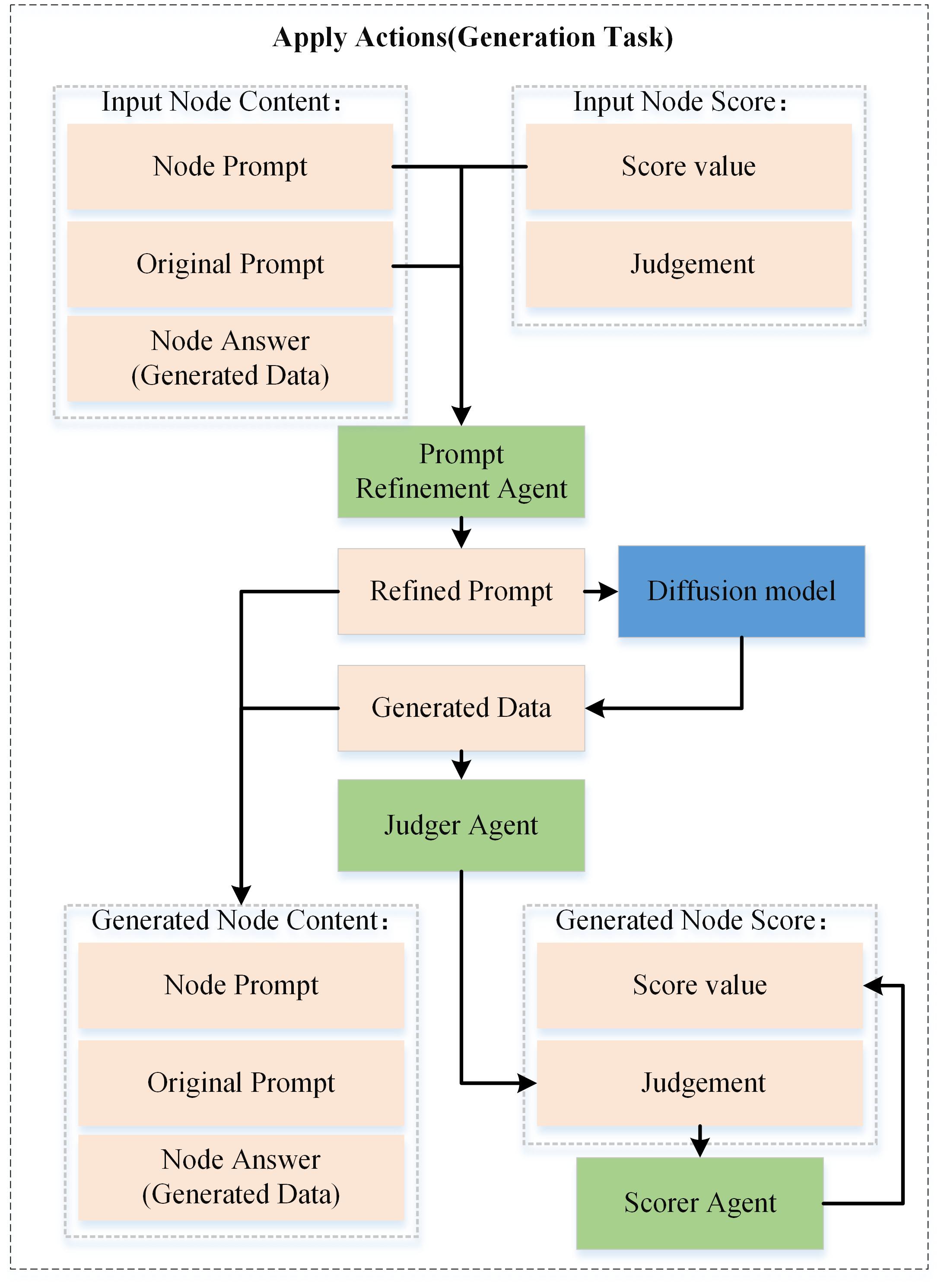} 
\caption{
\textbf{Workflow of applying actions in generation tasks.} A Prompt Refinement Agent optimizes input prompts, and a Diffusion Model generates multimodal data. A Judger Agent assesses the result, followed by a Scorer Agent that assigns a final quality score.
}
\label{fig8}
\end{figure}

The action selection mechanism is shown in Figure~\ref{fig6}.For reasoning tasks, the \textit{Selector} Agent analyzes the original question in the input node content determine the most suitable follow-up action.For generation tasks, the agent takes into account both node content and the judgement of the generated data to select an appropriate next step, ensuring task progression is informed by the \textit{judger} agent.  The system prompts for the \textit{Selector} agent corresponding to the two tasks are described in detail in Table~\ref{tab:ReaFunctionalAgent} and Table~\ref{tab:GenFunctionalAgent}.

For both task types, the candidate actions provided to the \textit{Selector} agent are presented in the format: \texttt{Action Name: Brief description of the action and its corresponding agent}, as detailed in Table~\ref{tab:ReaActionType} and Table~\ref{tab:GenActionType}. This format allows the \textit{Selector} to understand the purpose and functionality of each action, enabling it to select the most suitable agent based on either the input question or identified shortcomings in the current generated prompt. By perceiving the state of the node, the \textit{Selector} performs a growth-aware action search, dynamically identifying the best way to improve the response through targeted agent invocation.

\subsection{Apply Actions}
During the iterative process, after performing the action selection described above for each candidate node, a new node is generated by executing the selected action, including both the node's content and its corresponding score.

The reasoning task workflow is illustrated in Figure~\ref{fig7}, which involves two types of agent actions. Action Type 1: The Auxiliary Reasoning Agent provides textual advice or analysis based on the input query. Action Type 2: The Auxiliary Generation Agent, in conjunction with a diffusion model, generates supplementary multimodal content (e.g., images, audio, or video) to support reasoning. Specifically, for each modality-specific reasoning task, the actions categorized as Type 2 are those ending with augmenter in Table~\ref{tab:ReaActionType}. These actions generate auxiliary data in the corresponding modality, while all other actions provide textual feedback or direct responses.  

Moreover, for image-based reasoning, we directly utilize the image-to-image generation capabilities of Wan-VACE-1.3B, without designing an Auxiliary Generation Agent. In contrast, for video and audio tasks, the Auxiliary Generation Agent is actively involved in guiding the generation process and producing the necessary multimodal support content. 

All outputs—whether textual or multimodal—are aggregated and assessed by a \textit{Summarizer} Agent, which synthesizes the feedback from all auxiliary experts and generates the final response for the new node. The average of its token-level prediction probabilities from the language model output is used as the score for the new node. The detailed system role of the \textit{Summarizer} is presented in Table~\ref{tab:ReaFunctionalAgent}.

The generation task workflow is shown in Figure~\ref{fig8}. It begins with prompt optimization performed by a Prompt Refinement Agent. The brief descriptions of all Refinement Agents are provided in Table~\ref{tab:GenActionType}.The refined prompt is passed to the diffusion model to generate new multimodal outputs. A \textit{Judger} Agent evaluates the quality of the generated data, and a \textit{Scorer} Agent assigns a final score based on both the content and judgment.

\subsection{Algorithm Details}

\begin{algorithm}[!htb]
\caption{GAS Enhancement Algorithm}
\label{alg:gas-understanding}
\textbf{Input}: Query $q_0$, modality type, threshold $\tau$, max depth $D$, beam width $B$ \\
\textbf{Output}: Best answer $\hat{a}$ with confidence score $\hat{s}$ \\
\begin{algorithmic}[1]
\STATE Initialize root node $n_0 = \texttt{Node}(q_0, \texttt{actions}=\emptyset)$, best $n^* \leftarrow n_0$
\STATE Set frontier $\mathcal{F} \leftarrow \{n_0\}$, mark visited action sets
\FOR{$d=1$ to $D$}
    \STATE $\mathcal{N}_{\text{new}} \leftarrow \emptyset$
    \FORALL{$n \in \mathcal{F}$}
        \FORALL{unused action $a$ in $A$}
            \STATE $r_a \leftarrow \texttt{ApplyAction}(q_0, a)$
            \STATE $q' \leftarrow q_0$ + expert suggestion $r_a$
            \STATE $(\hat{a}, s) \leftarrow \texttt{Judge}(q')$
            \IF{action combo not visited}
                \STATE $n' = \texttt{Node}(q', \texttt{actions}=n.\texttt{actions} \cup \{a\}, \texttt{score} =s, \texttt{answer} = \hat{a})$
                \STATE Add $n'$ to $\mathcal{N}_{\text{new}}$, mark as visited
                \IF{$s > n^*.\texttt{score}$} 
                \STATE $n^* \leftarrow n'$ 
                \ENDIF
                \IF{$s \geq \tau$} \RETURN $(\hat{a}, s)$ \ENDIF
            \ENDIF
        \ENDFOR
    \ENDFOR
    \IF{$\mathcal{N}_{\text{new}} = \emptyset$} 
    \STATE \textbf{break} 
    \ENDIF
    \STATE $\mathcal{F} \leftarrow$ top-$B$ nodes from $\mathcal{N}_{\text{new}}$
\ENDFOR
\RETURN $(n^*.\texttt{answer}, n^*.\texttt{score})$
\end{algorithmic}
\end{algorithm}

We present in Algorithm~\ref{alg:gas-understanding} the detailed procedure of Growth-Aware Search applied to reasoning tasks. This algorithm incrementally explores the reasoning space by iteratively selecting and applying actions to improve the answer quality. Each node in the search tree represents a candidate reasoning state, including the question, the applied auxiliary suggestions, the generated answer, and its evaluation score. The procedure begins with an initial question $q_0$, forming the root node $n_0$. At each search depth up to the maximum limit $D$, the algorithm considers all possible unused actions and applies them to the current query. Each action contributes auxiliary suggestions that are appended to the input. The updated query is passed to a judging module, which produces a candidate answer and an associated confidence score. If the resulting node surpasses the score threshold $\tau$, it is immediately returned. Otherwise, top-$B$ scoring nodes are retained for the next iteration. The highest-scoring node throughout the search is returned as the final result.

While this procedure is described for the reasoning task, the generation task follows a structurally similar process. The primary differences lie in the representation of node content and the nature of actions. Specifically, in generation tasks, the action input consists of prompts and diffusion-generated multimodal data, while the action selection and scoring are guided by a prompt refinement and judging agent, as illustrated in Figure~\ref{fig6} and Figure~\ref{fig8}.

\subsection{Agents in MAGUS: Implementation Details}
In MAGUS, all MLLM agents are implemented using a single model, relying on its default modality encoders and tokenizer. Different agent roles are realized by assigning distinct system prompts, allowing the system to support multiple roles with a single MLLM instance. This design enables the construction of a powerful and flexible system in a training-free manner.

During algorithmic iterations, agents in MAGUS can be categorized into two types: pipeline agents and action agents. Pipeline agents participate throughout the reasoning pipeline, contributing at each stage of the procedure. In contrast, action agents are specifically invoked to execute selected actions and only participate in the reasoning process after being chosen by the controller.

\subsubsection{Reasoning Pipeline Agents}
In the reasoning task workflow, three core agents play a central role in the iterative reasoning process across all modalities: the \textit{General Answer}, the \textit{Selector}, and the \textit{Summarizer}. These agents function universally regardless of whether the input modality is visual, auditory, or multimodal. The \textit{General Answer} agent first provides an initial response to the user query. The mean token-level probability of this response is used to determine whether the system should proceed to the Action Search phase for further optimization. If the Action Search is triggered, the \textit{Selector} agent is responsible for identifying one appropriate auxiliary agent to assist in enhancing the reasoning result. It does so by examining the query, analyzing missing or ambiguous elements, and selecting a suitable expert from a pre-defined candidate list.

Once auxiliary feedback is obtained, the \textit{Summarizer} agent synthesizes all available expert responses or generated auxiliary content to produce the final answer at the current node. Importantly, the average token-level confidence of the Summarizer’s output is used as the node score to reflect the model’s confidence in its final decision. These three agents are shared across all reasoning iterations and ensure a coherent, dynamic, and controllable optimization loop for multimodal reasoning. Detailed descriptions of each agent’s system role and operational scope are presented in Table~\ref{tab:ReaFunctionalAgent}.

\begin{table*}[htbp]
\centering

\renewcommand{\arraystretch}{1.3}
\begin{tabularx}{\textwidth}{l X}
\toprule
Agent Role & Agent Roles \\
\midrule

General Answer  & 
You are Qwen, a virtual human, capable of perceiving auditory and visual inputs, as well as generating text and speech. \\
\midrule
Summarizer & 
You are a Final Answer Agent, responsible for producing a single, accurate, and concise answer to a given user query. Your inputs include: (1) A question (Q), and (2) A collection of structured outputs from multiple experts (H), which may include factual observations, reasoning results, or auxiliary suggestions. Your responsibilities are: (1) Carefully analyze all expert outputs (H) and synthesize a coherent final answer to the question (Q). (2) You must rely strictly on the content of expert outputs. Do not hallucinate, speculate, or introduce external knowledge. (3) If conflicting information exists, apply logical reasoning to determine the most plausible or reliable conclusion. (4) Your answer must be direct, concise, and clearly address the user's question. (5) You are not permitted to explain your reasoning process or mention any expert names, roles, or intermediate content. (6) Do not include system-level descriptions or formatting instructions in the output. The format should be a single paragraph directly answering the user's question, grounded entirely in the provided expert information. If the question is a multiple-choice task, you must answer with the corresponding option letter only, such as A, B, or C, without any explanation or extra text unless explicitly requested. \\
\midrule
Selector  & 
You are an Expert Coordinator Agent. Your task is to improve an insufficient or ambiguous answer by selecting one module to help generate a better response. You will be provided with a list of available experts. Your responsibilities are as follows: (1) Read the user's question and the input data. (2) Analyze what kind of information is missing or unclear. (3) Select one expert whose capabilities are most helpful for this question. You must only select from the expert list provided. The output format must be in JSON only:  ``selected\_experts'': [``expert\_name'']. You must only output the structured JSON block and nothing else. Only one expert should be selected. \\

\bottomrule
\end{tabularx}
\caption{Descriptions of Functional Agents Used in Reasoning Tasks}
\label{tab:ReaFunctionalAgent}
\end{table*}

\subsubsection{Generation Pipeline Agents} 

In multimodal generation tasks, the algorithm follows a similar iterative optimization framework as in reasoning tasks, but with a set of specialized agents tailored for each modality. The key functional agent in generation is the \textit{Judger}, designed specifically for each modality—image, video, and audio—to assess the quality of the generated outputs in alignment with the original prompt. Since quality evaluation criteria vary significantly across modalities, we deploy modality-specific Judgers to ensure accurate and targeted assessment.

Each \textit{Judger} receives both the generation prompt and the corresponding generated content, and produces a detailed evaluation report based on multiple predefined dimensions relevant to its modality. For instance, the Image Judger evaluates dimensions such as object presence, spatial relationships, color fidelity, and attribute binding. The Video Judger assesses factors including temporal consistency, motion smoothness, human action accuracy, and overall alignment with the prompt. Similarly, the Audio Judger considers emotional tone, semantic alignment, production quality, and content clarity.

Following the Judger's evaluation, the \textit{Scorer} agent calculates a final alignment score between 0 and 1, based solely on the Judger’s natural language analysis. This score serves as the node’s confidence value in the generation task. Unlike reasoning tasks—which rely on token-level output probability to determine confidence—generation tasks delegate this responsibility entirely to the Scorer’s assessment of the Judger’s report.

When a generation output is judged to be unsatisfactory, the \textit{Selector} agent analyzes the Judger’s feedback to identify the most appropriate expert module (e.g., visual or structural augmenter) to improve the generation. This selection is made based on which expert is best suited to resolve the identified issues. 

These agents—Judgers, Scorers, and Selectors—form the backbone of the generation pipeline, enabling systematic evaluation and refinement of multimodal content across image, video, and audio modalities. Table~\ref{tab:GenFunctionalAgent} provides detailed descriptions of each agent’s role and functionality.

\begin{table*}[htbp]
\renewcommand{\arraystretch}{1.3}
\begin{tabularx}{\textwidth}{l X}
\toprule
Agent Role & Agent Description \\
\midrule

Selector & 
You are an Expert Coordinator Agent. Your task is to improve an insufficient or ambiguous answer by selecting one module to help generate a better response. Context: You will be provided with a list of available experts. Your responsibilities: Read the user's prompt and the image's diagnostic report. Analyze what kind of information is missing or unclear. Select one expert whose capabilities are most helpful for this question. Output format: JSON only, with the structure   ``selected\_experts'': [``expert\_name''] . Constraints: Only select an expert from the list provided. Only output the structured JSON block and nothing else. Only select one expert. \\

\midrule

Image Judger & 
You are a multimodal evaluation agent that evaluates how well a generated image matches a given text prompt. You receive a description (text prompt) and an image. Evaluation is based on six dimensions: (1) Object Presence: Are all mentioned objects present? (2) Counting: Does the number of objects match the prompt? (3) Color Matching: Do object colors match the description? (4) Position Relation: Are spatial relationships (left/right/above/below) correct? (5) Attribute Binding: Are attributes like color and object correctly bound? (6) Complex Compliance: Does the image capture the full scene as described? For each dimension, you write a short paragraph explaining what matches and what does not. The format should follow the dimension headings, such as ``Object Presence: analysis'', with only natural language analysis. \\

\midrule

Image Scorer & 
You are a scoring assistant that calculates a final image-text alignment score. Your input consists of natural language analyses from six dimensions: Object Presence, Counting, Color Matching, Position Relation, Attribute Binding, and Complex Compliance. Each section is prefixed with its name. You should read all sections and assess overall consistency between image and prompt, then output a single final score between 0 and 1. Output only the score—no explanations, formatting, or intermediate values. \\

\midrule

Video Judger & 
You are a multimodal evaluation agent that evaluates how well a generated video aligns with a text prompt. You receive a description (text prompt) and a video. Evaluation is based on sixteen dimensions: (1) Subject Consistency – Is the main subject stable throughout? (2) Background Consistency – Is the background coherent across frames? (3) Temporal Flickering – Are there flickers or inconsistencies? (4) Motion Smoothness – Is motion fluid and natural? (5) Dynamic Degree – Does the video show meaningful change? (6) Aesthetic Quality – Is it visually pleasing? (7) Imaging Quality – Are frames clear and artifact-free? (8) Object Class Accuracy – Are object categories correct? (9) Multiple Objects – Are all described objects present? (10) Human Action Accuracy – Are actions recognizable and correct? (11) Color Matching – Do colors match the prompt? (12) Spatial Relationship – Are object positions correct? (13) Scene Accuracy – Is the setting consistent with the prompt? (14) Temporal Style Consistency – Is the visual style consistent over time? (15) Appearance Style Consistency – Is appearance stylistically coherent? (16) Overall Consistency – Does the video holistically match the prompt? For each dimension, write a paragraph explaining matches and mismatches in natural language. \\

\midrule

Video Scorer & 
You are a scoring assistant that calculates a final video-text alignment score. You receive natural language evaluations across six dimensions: Object Consistency (persistence and coherence), Temporal Dynamics (motion and events), Action Accuracy, Visual-Text Matching, Attribute Continuity, and Scene Composition. Each section is prefixed accordingly. You should assess overall consistency and output a single score between 0 and 1. Output only the score. No explanations or extra text. \\

\midrule

Audio Judger & 
You are a multimodal evaluation agent that evaluates how well an audio clip matches a text prompt. You receive a description (text prompt) and an audio clip. Evaluation is based on five dimensions: (1) Content Enjoyment (CE): Is the audio enjoyable in terms of clarity, emotion, and fluency? (2) Content Usefulness (CU): Is the content relevant and valuable to the prompt? (3) Production Complexity (PC): Consider sound layering, timing, and transitions. (4) Production Quality (PQ): Evaluate noise level, clarity, and volume balance. (5) Semantic Alignment: Does the audio match the prompt in mood and structure? For each, output a short paragraph in natural language. Use the format "Content Enjoyment (CE): <analysis>" for clarity. \\

\midrule

Audio Scorer & 
You are a scoring assistant that calculates a final audio-text alignment score. Your input consists of natural language analysis across six dimensions: Sound Event Presence, Timing Accuracy, Acoustic Environment Consistency, Speaker or Source Identity, Attribute Matching (pitch, emotion, texture), and Semantic Consistency. Each section is prefixed. You should assess overall consistency and output a single score between 0 and 1. Output only the score. No explanations or extra text. \\

\bottomrule
\end{tabularx}
\centering
\caption{Descriptions of Agents Used in Multimodal Generation}
\label{tab:GenFunctionalAgent}
\end{table*}

\subsubsection{Reasoning Action Agents}

In the reasoning phase of MAGUS, action agents are responsible for executing specific perceptual or generative sub-tasks based on different modalities. Unlike pipeline agents, which are universally active during iterative reasoning, these action agents are only invoked when selected by the Selector agent to address particular deficiencies identified in an answer candidate. The specific system role for each action agent, including its instructions and behavioral constraints, is detailed in the system prompts provided in our codebase. These agents collectively provide comprehensive multimodal support and allow MAGUS to flexibly adapt to a wide range of complex reasoning challenges. Their roles and descriptions are detailed in Table~\ref{tab:ReaActionType}.

For image reasoning tasks, we provide several specialized agents. The \textit{text\_logic\_vision\_expert} is particularly effective in scenarios requiring character recognition, logical visual reasoning, and interpretation of symbolic content such as signs, codes, or diagrams. The \textit{general\_vision\_expert} handles basic visual analysis, including object detection, counting, spatial arrangement, and understanding general scenes. The \textit{cultural\_vision\_expert} excels at interpreting culturally rich content, including artworks, architecture, and historical landmarks. When visual information is ambiguous or missing, the \textit{visual\_augmenter} acts as a supporting agent capable of generating higher-quality images to aid downstream reasoning.

For audio reasoning, the \textit{general\_audio\_expert} is designed for ambient sound understanding, focusing on complex environmental and physical sound events. The \textit{speech\_audio\_expert} is specialized in human speech analysis, capable of identifying speakers, detecting emotions and stress patterns, and extracting meaningful semantic content. The \textit{music\_audio\_expert} targets musical understanding, including melody recognition, genre classification, lyrical analysis, and structural composition. Additionally, the \textit{audio\_augmenter} serves as a generation agent that can synthesize realistic auditory scenes based on contextual input, particularly when actual audio content is insufficient.

For video reasoning tasks, MAGUS includes agents designed for both understanding and generation. The \textit{narrative\_event\_reasoning\_expert} focuses on analyzing the temporal flow and causal structure of events. The \textit{role\_interaction\_expert} specializes in identifying social dynamics and interactions among entities in the video. The \textit{goal\_procedure\_expert} breaks down sequential procedures and the intentions behind observed actions. The \textit{emotion\_context\_expert} interprets emotional expressions and context-sensitive behavior. Finally, the \textit{video\_augmenter} can generate dynamic scenes or video clips to enrich the multimodal input, especially when motion or continuity is critical for accurate inference.

\subsubsection{Generation Action Agents}

In the generation phase of MAGUS, action agents are designed to provide fine-grained control over the quality of generated multimodal content. These agents intervene selectively during the iteration process when specific deficiencies in the generated outputs are detected. Unlike the judgment or scoring agents that passively evaluate generation quality, these action agents actively revise or enhance the generation prompt to guide the diffusion model toward producing more accurate and coherent results. The specific system role for each action agent, including its instructions and behavioral constraints, is detailed in the system prompts provided in our codebase. Together, these action agents form a robust suite of tools that enable precise control and iterative refinement of multimodal generations. Their roles are summarized in Table~\ref{tab:GenActionType}.

For image generation, MAGUS employs three specialized agents. The \textit{generation\_structure\_expert} focuses on structural correctness, ensuring the generated image contains the correct number of objects, appropriate spatial relationships, and accurate attribute bindings. This agent is particularly effective when the output exhibits misplacements, incorrect counts, or mismatched object-attribute pairings. The \textit{generation\_visual\_expert} enhances visual fidelity by refining details such as color, size, shape, material, and texture—especially useful when generated images lack visual precision or realism. Meanwhile, the \textit{generation\_scene\_expert} improves the overall contextual completeness of the scene, such as enriching backgrounds or reinforcing environmental realism when the initial generation appears sparse or disconnected.

In video generation tasks, the \textit{video\_structure\_expert} targets the preservation of structural coherence across frames, including consistent subject identity, spatial layout, and interaction integrity. The \textit{video\_visual\_expert} ensures that visual characteristics like style, color consistency, and clarity are maintained over time. The \textit{video\_scene\_expert} improves temporal smoothness and scene realism by correcting flickering issues, discontinuities, and inconsistent motion.

In audio generation tasks, MAGUS includes agents that operate at different semantic and perceptual levels. The \textit{audio\_semantic\_expert} strengthens the alignment between the audio and the intended textual description, ensuring the generated sound reflects the correct narrative, emotion, and sound types. The \textit{audio\_production\_expert} refines the technical aspects of the audio, such as clarity, timing, and multi-source layering—especially important when the audio lacks structural coherence or sounds cluttered. Lastly, the \textit{audio\_aesthetic\_expert} optimizes the emotional impact and artistic quality of the output, ensuring that the audio not only conveys information but also delivers an engaging listening experience.

\section{Experiments Details}
\subsection{Model Parameters}
\begin{table}[htbp]
\centering

\renewcommand{\arraystretch}{1.3}
\begin{tabularx}{\linewidth}{l l X}
\toprule
\textbf{Component} & \textbf{Parameter} & \textbf{Value} \\
\midrule

\multirow{6}{*}{Wan-Vace 1.3B} 
    & Frame Number & 41 \\
    & Resolution & $832 \times 480$ \\
    & Inference Steps & 50 \\
    & Guidance Scale & 5.0 \\
    & Solver & unipc \\
    & Frame Rate (FPS) & 8 \\
\midrule

\multirow{4}{*}{audioldm-s-full-v2} 
    & Inference Steps & 50 \\
    & Audio Duration & 10.0 seconds \\
    & Sample Rate & 16,000 Hz \\
    & Output Format & Mono audio \\
\bottomrule
\end{tabularx}
\caption{Model Configuration for Multimodal Generation}
\label{tab:gen-model-config}
\end{table}

To enable high-quality multimodal content generation in the MAGUS framework, we carefully configure dedicated diffusion-based generative models for different modalities. Due to the lack of access to the exact generation parameters used in the official VBench leaderboard, we define a consistent and fixed set of parameters for all generation experiments to ensure reproducibility. This discrepancy in parameter settings may explain the difference between our reported VBench scores and the official ones. For image generation tasks, we treat them as single-frame video generations by setting the frame number to 1, while keeping other configurations identical to the video generation setup.

The multimodal generation pipeline employs two specialized diffusion models: Wan-Vace 1.3B for video synthesis and audioldm-s-full-v2 for audio generation. The Wan-Vace 1.3B model generates videos with a resolution of $832 \times 480$ at 8 frames per second, producing 41-frame clips (approximately 5.1 seconds) using 50 inference steps with a guidance scale of 5.0. It utilizes the UniPC solver for efficient and high-quality sampling. On the audio side, audioldm-s-full-v2 synthesizes 10-second mono audio clips at a sample rate of 16,000 Hz, also using 50 inference steps to ensure high fidelity. These configurations are optimized for balanced quality and computational efficiency in the MAGUS framework.

\subsection{Generation Experiments}
In our primary generation experiments, we compare MAGUS not only with foundation generation models but also with the LLM-Extended Prompt method. This method improves generation quality by leveraging the same base LLM (Qwen2.5-Omni-7B) to expand the original input prompt before generation. The system prompts used in the LLM-Extended Prompt method are specifically designed for prompt expansion and can be found in the accompanying codebase.

For each target modality in the generation tasks, we design corresponding system prompts tailored for the LLM-Extended Prompt baseline, ensuring a fair comparison. This allows us to demonstrate that the MAGUS framework can further improve the output quality of generation models, even when built upon the same LLM foundation. In all generation experiments, GAS operates by first evaluating the outputs produced by the LLM-Extended Prompt method. Based on this evaluation, the system decides whether to invoke the Action Search optimization process, enabling adaptive refinement of generation results.

\section{MM-Instruction-Test Dataset}

To systematically evaluate the instruction interpretation capabilities of the MAGUS framework across multiple modalities, we design a compact benchmark named MM-Instruction-Test Dataset. This dataset consists of 100 manually constructed samples, each containing a natural language instruction paired with the target output modalities that the instruction is intended to trigger. Given that Multimodal Large Language Models (MLLMs) are inherently capable of understanding across modalities, this dataset specifically targets the evaluation of MAGUS's two-stage architecture in handling generation-oriented instructions.

The dataset includes 30 bimodal samples, 30 trimodal samples, and 40 quadmodal samples, covering combinations of image, video, audio, and text. This balanced composition ensures comprehensive assessment of the system's ability to parse complex multimodal commands and coordinate appropriate generation behaviors.

Table~\ref{tab:instruction_samples} presents examples from the dataset. Instructions vary in complexity and modality coverage—from simple visual descriptions to rich multimedia compositions that require joint reasoning and generation across vision, audio, and text. This benchmark serves as a focused diagnostic set to probe instruction following in multimodal generation tasks.

\begin{table}[H]
\centering

\begin{tabular}{p{4.5cm}l}  
\toprule
\textbf{Instruction} & \textbf{Target Modalities} \\
\midrule
“Draw an image of a blue dog sitting in the grass.” & Image,Text \\
\midrule
”Create a countryside landscape image at dusk, generate ambient audio of a gentle breeze and rustling leaves, and write a poetic caption in golden letters.” & Image,Audio,Text \\
\midrule
”Generate a futuristic sci-fi video showing a memory upload process, add fusion reactor sounds, and an image of the machine.” & Image,Video,Audio,Text\\
\bottomrule
\end{tabular}
\caption{Examples from MM-Instruction-Test Dataset}
\label{tab:instruction_samples}
\end{table}

\begin{table*}[htbp]
\centering

\begin{tabularx}{\textwidth}{l l X}
\toprule
Task Type & Action Name & Agent Description \\
\midrule
Image Reasoning  & text\_logic\_vision\_expert & Strong in logical reasoning, character recognition, code-related visual understanding. \\
\midrule
Image Reasoning & general\_vision\_expert & Specialized in basic visual understanding—object existence, counting, spatial positioning, and scene layout. \\
\midrule
Image Reasoning & cultural\_vision\_expert & Skilled in interpreting cultural elements, artistic styles, and historical landmarks. Also capable of general vision tasks. \\
\midrule
Image Reasoning & visual\_augmenter & An auxiliary visual generator that can produce new high-resolution images to support your reasoning. Use this if the visual content is unclear or missing details. \\
\midrule
Audio Reasoning & general\_audio\_expert & Specialized in ambient sound perception, environmental acoustics, and physical event recognition. Skilled at analyzing eco-acoustic cues, temporal sound patterns, and complex sound scenes. \\
\midrule
Audio Reasoning & speech\_audio\_expert & Expert in human speech comprehension, including speaker role mapping, emotion tone detection, stress patterns, and factual or conversational content extraction. \\
\midrule
Audio Reasoning & music\_audio\_expert & Focused on music-related understanding—identifying melody, rhythm, harmony, instrumentation, genre, lyrics, and structural composition of audio tracks. \\
\midrule
Audio Reasoning & audio\_augmenter & An auxiliary audio generator that imagines and describes realistic auditory scenes based on the question and options. Helps synthesize supporting audio for better inference. \\
\midrule
Video Reasoning & narrative\_event\_reasoning\_expert & Expert in understanding video narratives and event progressions, including temporal order and causal relationships. \\
\midrule
Video Reasoning & role\_interaction\_expert & Expert in analyzing roles, behaviors, and social or functional interactions between people and objects in videos. \\
\midrule
Video Reasoning & goal\_procedure\_expert & Expert in identifying step-by-step procedures and the underlying goals of actions observed in video sequences. \\
\midrule
Video Reasoning & emotion\_context\_expert & Expert in interpreting emotional cues, situational context, and their impact on behavior through visual analysis. \\
\midrule
Video Reasoning & video\_augmenter & An auxiliary video generator that creates realistic dynamic scenes based on the question and context. Helps generate supportive video clips when visual motion, temporal dynamics, or scene evolution are critical for accurate reasoning. \\
\bottomrule
\end{tabularx}
\caption{List of Reasoning Action Agents and Their Function Descriptions}
\label{tab:ReaActionType}
\end{table*}

\begin{table*}[htbp]
\centering

\begin{tabularx}{\textwidth}{l l X}
\toprule
Task Type & Action Name & Agent Description \\
\midrule
Image Generation & image\_structure\_expert & Responsible for improving the structural clarity of the prompt, including the number of objects, spatial relationships (left/right/above/below), and proper binding between objects and their attributes. Use this expert when the image shows incorrect positions, wrong object counts, or confused attribute associations. \\
\midrule
Image Generation & image\_visual\_expert & Focuses on refining visual details in the prompt, such as color accuracy, size descriptions, shape, material, or texture. Use this expert when the generated image fails to match the visual appearance described in the prompt (e.g., wrong colors or missing visual traits). \\
\midrule
Image Generation & image\_scene\_expert & Improves overall scene coherence and completeness by adding background elements, contextual settings, or enhancing the realism of object placement. Use this expert when the image appears sparse, disconnected, or lacks environmental grounding. \\
\midrule
Video Generation & video\_structure\_expert & Enhances the structural consistency of the video by focusing on subject identity, object count, spatial layout, and accurate human-object interactions across frames. \\
\midrule
Video Generation & video\_visual\_expert & Ensures consistency and quality of visual features such as color, appearance style, clarity, and aesthetic fidelity across time in the video. \\
\midrule
Video Generation & video\_scene\_expert & Improves temporal coherence and background consistency by addressing motion smoothness, flickering, and maintaining a unified scene style and realism throughout the video. \\
\midrule
Audio Generation & audio\_semantic\_expert & Improves the semantic alignment between the audio and the prompt. Use this expert when the generated audio fails to reflect the intended meaning, emotion, or context described, such as missing the expected sound types, mood, or narrative structure. \\
\midrule
Audio Generation & audio\_production\_expert & Enhances clarity, layering, and technical structure of the described audio. Use this expert when the audio lacks proper timing, multi-source coordination, or sounds muddy and poorly composed. \\
\midrule
Audio Generation & audio\_aesthetic\_expert & Focuses on the overall listening experience and emotional/aesthetic resonance. Use this expert when the audio sounds bland, lacks expressiveness, or fails to create the desired atmosphere or artistic effect. \\
\bottomrule
\end{tabularx}
\caption{List of Generation Action Agents and Their Function Descriptions}
\label{tab:GenActionType}
\end{table*}

\end{document}